\documentclass[letterpaper,10pt, conference]{_Aux/IEEEconf}
\IEEEoverridecommandlockouts
\usepackage{times}
\usepackage{setspace}
\usepackage{url}
\spacing{1}
\usepackage[utf8]{inputenc}
\usepackage[T1]{fontenc}
\usepackage{graphicx}		
\usepackage{wrapfig}
\usepackage[format=plain,font=footnotesize,labelfont=bf,labelsep=period]{caption}
\usepackage{sidecap} 
\usepackage{subfig}
\usepackage[export]{adjustbox}
\usepackage[font=small]{caption}
\usepackage{float}
\usepackage{scalerel}

\usepackage{amsmath} 
\usepackage{amssymb}  
\usepackage{amsthm}
\usepackage{mathtools}
\usepackage[normalem]{ulem}
\usepackage{paralist}	
\usepackage[space]{grffile} 
\usepackage{color}
\let\labelindent\relax
\usepackage{enumitem}
\usepackage{bm}
\usepackage{cancel}
\usepackage{hhline}
\usepackage[c2 , nocomma]{optidef}

\usepackage{hyperref}

\DeclareMathOperator*{\argmax}{arg\,max}
\newtheorem{theorem}{Theorem}

\theoremstyle{definition}
\newtheorem{definition}{Definition}
\theoremstyle{remark}

\theoremstyle{definition}

\theoremstyle{definition}

\newcommand{\R}{\mathbb{R}}
\newcommand{\C}{\mathcal{C}}

\newcommand{\K}{\mathcal{K}}

\definecolor{blue}{RGB}{38,38,134}
\definecolor{darkblue}{RGB}{0,0,102}
\definecolor{lightblue}{RGB}{77,77,148}

\definecolor{gold}{RGB}{234, 170, 0}
\definecolor{metallic_gold}{RGB}{139, 111, 78}

\DeclareMathOperator*{\argmin}{arg\,min}

\usepackage{xfrac}

\usepackage{dblfloatfix}

\usepackage{cite}
\begin{document}

 \title{ \bf
 A Learning-Based Framework for Safe Human-Robot Collaboration with Multiple Backup Control Barrier Functions }

 \author{Neil C. Janwani$^{1*}$, Ersin Da\c{s}$^{2*}$,  Thomas Touma$^{2}$, Skylar X. Wei$^{2}$, Tamas G. Molnar$^{3}$, Joel W. Burdick$^{2}$%
\thanks{*Both authors contributed equally.}
\thanks{**This work was supported by DARPA under the LINC program.}
\thanks{$^{1}$N. C. Janwani is with the Department of Computing and Mathematical Sciences, California Institute of Technology, Pasadena, CA 91125. ${\tt\small njanwani@caltech.edu}$}%
\thanks{$^{2}$E. Da\c{s}, S. X. Wei, T. Thouma, and J. W. Burdick are with the Department of Mechanical and Civil Engineering, California Institute of
Technology, Pasadena, CA 91125, USA. ${\tt\small \{ersindas, ttouma, swei, jburdick \}@caltech.edu}$ }
\thanks{$^{3}$T. Molnar is with the Department of Mechanical Engineering, Wichita State University, Wichita, KS 67260, USA. ${\tt\small tamas.molnar@wichita.edu}$}
}

\maketitle
\pagestyle{plain}

\begin{abstract}
Ensuring robot safety in complex environments is a difficult task due to actuation limits, such as torque bounds. This paper presents a safety-critical control framework that leverages learning-based switching between multiple backup controllers to formally guarantee safety under bounded control inputs while satisfying driver intention. By leveraging {\em backup controllers} designed to uphold safety and input constraints, \textit{backup control barrier functions} (BCBFs) construct implicitly defined control invariant sets via a feasible quadratic program (QP). However, BCBF performance largely depends on the design and conservativeness of the chosen backup controller, especially in our setting of human-driven vehicles in complex, e.g, off-road, conditions. While conservativeness can be reduced by using multiple backup controllers, determining when to switch is an open problem. Consequently, we develop a broadcast scheme that estimates driver intention and integrates BCBFs with multiple backup strategies for human-robot interaction. An LSTM classifier uses data inputs from the robot, human, and safety algorithms to continually choose a backup controller in real-time. We demonstrate our method's efficacy on a dual-track robot in obstacle avoidance scenarios. Our framework guarantees robot safety while adhering to driver intention.
\end{abstract}

\section{Introduction} \label{sec:intro}
\begin{figure}
    \centering
    \includegraphics[width=8cm]{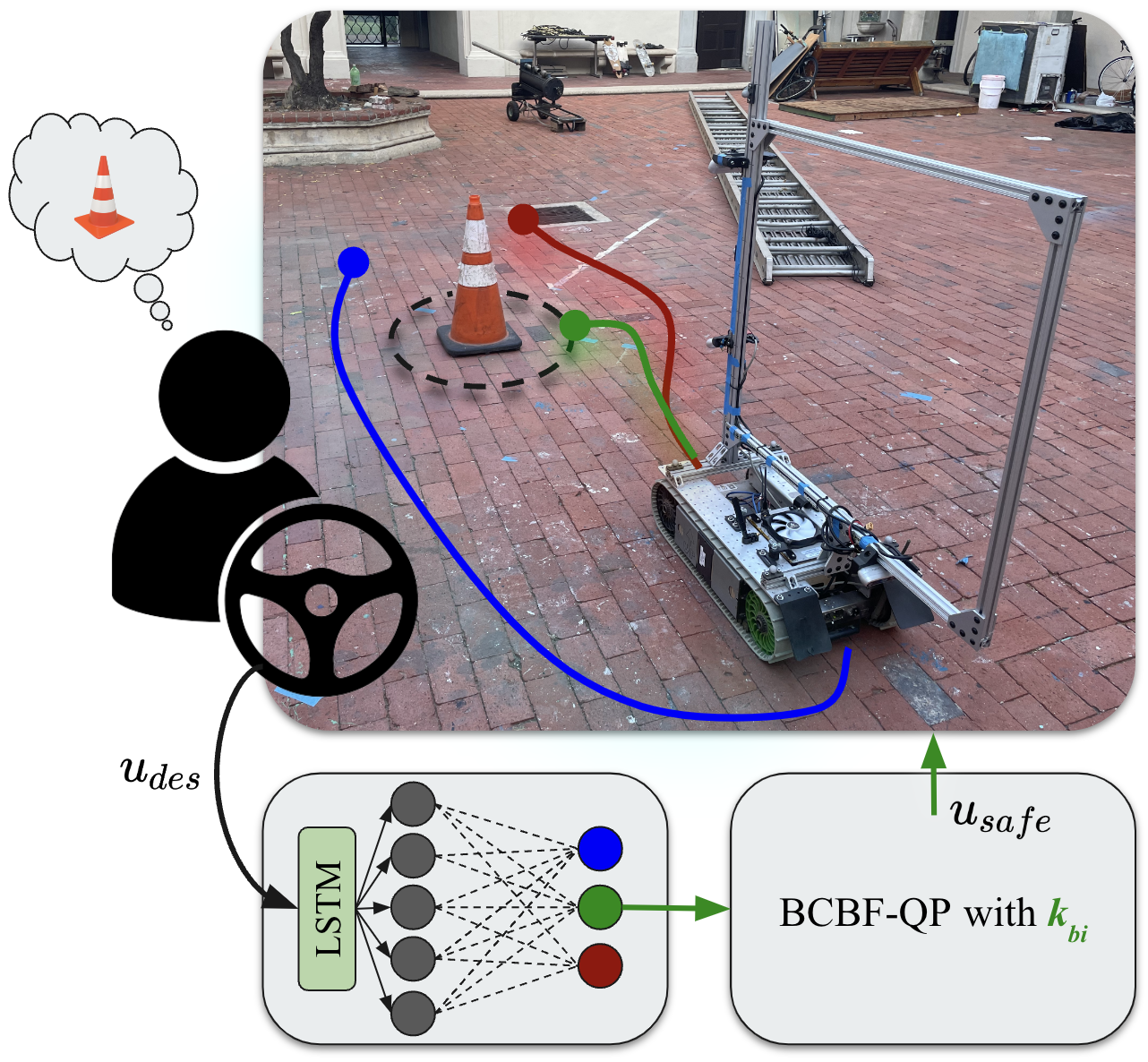}
    \caption{Visualization of the proposed safety-critical control framework with desired robot behavior. The red, green, and blue arrows represent different trajectories that rely on different backup controllers, with corresponding colors, to guarantee safety. Using driver intention tracking, the correct, green, backup controller is chosen among the red and blue controllers in order to guide the driver to their desired location. A supplementary video can be found here: \url{https://youtu.be/41Jh1GD_9Ok}}
    \label{fig:master}
    \vspace{-2em}
\end{figure}
\textit{Safety filters} are useful control tools that allow a robot to remain safe while under actuation from a potentially unsafe controller, or driver. Safety filters accomplish this by minimally affecting desired control commands, and thus have become a popular add-on to robot control architectures \cite{hobbs2023runtime,hsu2023safety,wabersich2021} since they address real-world robot dynamics and kinodynamic constraints in a run-time fashion.

{\em Control barrier functions} (CBFs) \cite{ames2017control} are a popular method for constructing safety filters due to their ability to integrate nonlinear dynamics while providing formal safety guarantees.  A CBF-based safety filter requires defining a control-invariant set that ensures safety.  However, such sets can be difficult to construct with input constraints in mind \cite{chen2021backup, breeden2023robust, agrawal2021safe, cortez2020correct,xiao2022sufficient,wiltz2023construction,folkestad2020data}.

Consequently, the CBF framework has been extended to include actuation capability, such as torque limits, through the use of {\em backup} control barrier functions (BCBFs) \cite{gurriet2020, molnar2023safety, cosner2021}.  BCBFs rely on the formulation of a {\em backup controller},
which is designed with input limits in mind to guarantee safety.

The backup controller
typically involves a simple saving maneuver, such as coming to a stop, turning at a maximum rate, or hovering.  By calculating the future backup trajectory of the system, one can analyze future safety of the robot and incorporate this information into optimization-based controllers like quadratic programs (QPs).  Consequently, when constructing safety filters using the BCBF method, the system's conservativeness is a strong function of the control engineer's choice of
backup policy.

To address this limitation, recent work examines the use of multiple backup strategies in the BCBF framework, since multiple strategies can help overcome the conservativeness of a single strategy. In \cite{chen2021obstacle}, an algorithm is used to evaluate the BCBF method with multiple backup controllers, such that the one with the least control intervention can be chosen. However, with many backup controllers, this method could be computationally infeasible. In \cite{singletary2022time-varying} and \cite{singletary2022}, different maneuvers are proposed to increase the reachability of a given backup controller, where a switching algorithm chooses a different maneuver if it is no longer possible to perform the current maneuver. While this method used multiple backup maneuvers with more computationally efficient ways of solving BCBFs, it is better suited to fully autonomous systems. It does not explicitly provide, in a human-robot interaction context, a way to take into account the intentions of a human vehicle driver or human companion.  Nor does it account for the fact that the human operator may have better situational awareness than the autonomous controller. Even though autonomous selection may be suboptimal at times, the backup control strategy is often fixed and independent of the driver's preferred nature of safety-critical behavior.   Generally, the types of safe behavior utilized by the BCBF cannot be tuned, leading to situations where potentially preferred safe behaviors are filtered out by the BCBF framework.

There has been impressive previous work to include human preferences in safety filters. For instance in \cite{cosner2022safety}, preference-based learning is used to adapt tuneable parameters of a CBF to user preferences for autonomous obstacle avoidance on a quadruped. However, input bounds are not formally considered, which could potentially result in infeasibility and consequently unsafe trajectories. While there has also been work on preference based reinforcement learning (pbRL) \cite{Novoseller_PBRL}, these methods must sample trajectories within the expected environment \cite{witrthRLSurvey}. Since the drivers of teleoperated systems must provide control inputs at each timestep, trajectory sampling becomes difficult to complete due to the human-in-the-loop nature of the problem.

In order to intelligently pick backup controllers, we propose a system that utilizes a long short term memory (LSTM) and a deep neural network (DNN) component to learn a reward corresponding to each backup controller. LSTMs have been used in real-time driver intention and maneuver tracking, and have shown to be successful at regression and classification tasks with trajectory input \cite{KhairdoostLSTM, WangBrakingLSTM,ZynerIntentionLSTM}. A simple switching law is derived by using the controller with the largest reward in the BCBF framework. This system learns to estimate driver intention by training on example trajectories, labeled with the correct choice of backup controller, as provided by the driver. In our work, we show that
\begin{enumerate}[label=\arabic*.]
    \item The LSTM-DNN architecture effectively learns a switching law for the BCBF framework with multiple backup controllers--choosing the controller closest to the driver's preference.
    \item Our approach increases the reachable sets of the driven robot while maintaining the safety guarantees of BCBFs.
\end{enumerate}
Moreover, we present experimental implementations of our algorithms on a tracked robot. These experiments demonstrate that switching between new and previously formulated backup controllers based on estimated driver intention can work on actual hardware systems.

This paper is organized as follows. Section \ref{sec:preliminaries} presents preliminaries on CBFs, BCBFs, and DNNs. Section \ref{sec:theory} presents our contributions, including a description of system architecture. We present the implementation of our framework on hardware in Section \ref{sec:implementation} and discuss experimental results in Section \ref{sec:results}. Section \ref{sec:future} concludes the paper. 

\section{Preliminaries} \label{sec:preliminaries}
We consider robots governed by a general nonlinear control affine system:
\begin{equation}
\label{system}
    \dot{x}  = f(x) + g(x) u,
\end{equation}
where ${x \in \mathbb{R}^n}$ is the state, ${u \in U \subset \mathbb{R}^m}$ is the control input, and functions ${f: \mathbb{R}^n \rightarrow \mathbb{R}^n}$, and ${g: \mathbb{R}^n \rightarrow \mathbb{R}^{n \times m} }$ are locally Lipschitz continuous. A locally Lipschitz continuous controller ${\mathbf{k}: \mathbb{R}^n \to U}$ yields a locally Lipschitz continuous closed-loop control system, ${f_{\rm cl}: X \to \mathbb{R}^n}$:
\begin{equation}
\label{eq:clsystem1}
        \dot{x} = {f}(x) + {g}(x) \mathbf{k}(x) \triangleq f_{\rm cl}(x).
\end{equation}
Given an initial condition ${x_0 \triangleq x(t_0)   \in \R^n}$, this system has a unique solution given by the flow map
\begin{equation}
\label{eq:flow1}
    \phi(t, x_0) \triangleq {x}(t) = x_0 + \int^{t}_{t_0} f_{\rm cl}(x(\tau)) d\tau ,~t > t_0.
\end{equation}

\subsection{Control Barrier Functions}
To characterize safety, we consider a safe set ${\mathcal{C} \subset \mathbb{R}^n}$ defined as the 0-superlevel set of a continuously differentiable function ${h: \mathbb{R}^n \rightarrow \mathbb{R}}$:
\begin{align}
\begin{split}
\label{eq:CBFset1}
    \mathcal{C} \triangleq \left\{ x \in X \subset \mathbb{R}^n : h(x) \geq 0 \right\}, \\
    \partial \mathcal{C} \triangleq \left\{ {x \in X \subset \mathbb{R}^n} : h(x) = 0 \right\},
\end{split}
\end{align}
where ${\partial \mathcal{C}}$ is the boundary of set $\mathcal{C}$. This set is forward invariant if, for every initial condition ${x(0) \in \mathcal{C}}$, the solution of \eqref{eq:clsystem1} satisfies ${x(t) \in \mathcal{C}, ~\forall t \geq 0}$. The closed-loop system \eqref{eq:clsystem1} is called safe w.r.t.~set $\mathcal{C}$ if $\mathcal{C}$ is forward invariant \cite{ames2017control}.

\begin{definition}[Control barrier function \cite{ames2017control}]
Function $h$ is a CBF for (\ref{system}) on $\mathcal{C}$ if ${\frac{\partial h}{\partial x} \neq 0}$ for all ${ x \in \partial \mathcal{C}}$ and there exists an extended class-$\mathcal{K}_{\infty}$ function\footnote{A continuous function $\alpha : \mathbb{R} \rightarrow \mathbb{R}$ belongs to the set of extended class-$\mathcal{K}$ functions ($\alpha \in \mathcal{K}_{\infty, e}$) if it is strictly monotonically increasing, $\alpha(0) = 0$, $\alpha(r) \rightarrow \infty$ as $r \rightarrow \infty$, $\alpha(r) \rightarrow -\infty$ as $r \rightarrow -\infty$.} $\alpha \in \mathcal{K}_{\infty, e}$ such that $\forall x \in \mathcal{C}$:
\begin{align}
\label{cbf}
   \sup_{u \in U}\! \Big [ \!{\dot{h}(x, u)}  \!=\!
    \dfrac{\partial h(x)}{\partial x}\! f(x)  \!+  \dfrac{\partial h(x)}{\partial x}g(x)  \!u \!\Big ] 
   \!\geq\! -\alpha (h(x)).
\end{align}
\end{definition}
\begin{theorem}\cite{ames2017control}
    \label{teo:cbfdef}
    If $h$ is a CBF for \eqref{system} on $\mathcal{C}$, then any locally Lipschitz continuous controller ${\mathbf{k}: \mathbb{R}^n \to U}$ satisfying 
    \begin{equation}
        \label{eq:cbf_def}
        \dot{h}\left(x, \mathbf{k}(x)\right) \geq - \alpha (h(x)),~~\forall x \in \mathcal{C}
    \end{equation}
    renders \eqref{eq:clsystem1} safe with respect to $\mathcal{C}$. 
\end{theorem}

\subsection{Backup Control Barrier Functions}
BCBFs are motivated by the fact that finding a function $h$ satisfying the CBF condition \eqref{cbf}, which is required for the feasibility of~\eqref{eq:cbf_def}, may be challenging for a particular choice of ${h}$, especially with bounded inputs.
Consider input bounds with component-wise hard constraints: 
\begin{align}
    \label{eq:umax}
    U \triangleq \left\{ u \in  \mathbb{R}^m : -u_{\rm max} \leq u \leq  u_{\rm max} \right\},
\end{align}
where ${u_{\rm max} \in \mathbb{R}^m_{> 0}}$.

The backup set method \cite{gurriet2020, molnar2023safety, cosner2021} addresses this feasibility problem by designing implicit control invariant sets and safe controllers through a CBF framework.

We consider a backup set ${\mathcal{C}_{\rm b} \!\subset\! \mathbb{R}^n }$ defined as the 0-superlevel set of a smooth ${h_{\rm b}: \mathbb{R}^n \!\to\! \mathbb{R}}$:
\begin{align}
\label{CBF1}
    \mathcal{C}_{\rm b} \triangleq \left\{ x \in  \mathbb{R}^n : h_{\rm b}(x) \geq 0 \right\}, 
\end{align}
such that it is a subset of ${\mathcal{C}}$, i.e., ${\mathcal{C}_{\rm b} \subseteq \mathcal{C}}$, and ${\frac{\partial h_{\rm b}}{\partial x} \neq 0}$ for all ${x \in \partial \C_{\rm b}}$.  Practically speaking, the backup set and backup controller are easily characterized safety procedures that can keep the vehicle in a strict subset of the maximally feasible safe set, which may be impossible to characterize in practice.

We assume that there is a locally Lipschitz continuous backup controller ${\mathbf{k_{b}}: \mathbb{R}^n \to U}$, which satisfies \eqref{eq:umax} for all ${x \in \mathcal{C}}$, to render the backup set forward invariant.
This results in the locally Lipschitz continuous closed-loop
\begin{equation}
\label{eq:clsystem}
       \dot{x} = {f}(x) + {g}(x) \mathbf{k_{b}}(x) \triangleq f_{\rm b}(x) ,
\end{equation}
and its solution with the initial state ${x_0 \in \mathbb{R}^n}$ is 
\begin{equation}
\label{eq:flow}
    \phi_{\rm b}(t, x_0) \triangleq {x}(t) = x_0 + \scaleobj{.9}{\int^{t}_{t_0}} f_{\rm b}(x(\tau)) d\tau ,\quad t > t_0 .
\end{equation}

Designing a control invariant backup set ${\mathcal{C}_{\rm b}}$ is generally easier than verifying if ${\mathcal{C}}$ is control invariant. However, the methods used to develop ${\mathcal{C}_{\rm b}}$ may result in a conservative set \cite{chen2021backup}. We can reduce conservatism by expanding the backup set. To achieve this, we use the backup trajectory ${\phi_{\rm b}(\tau,x)}$ over a finite time period ${\tau \in [0,T]}$ with some ${T \in \mathbb{R}_{> 0}}$. 

Note that ${{\phi_{\rm b}}(\tau, x)}$ is the flow map of the system under the backup controller with the initial condition ${x}$. We define a larger control invariant set, called $\mathcal{C}_{E}$, such that ${\mathcal{C}_{\rm b} \!\subseteq\! \mathcal{C}_{E} \!\subseteq\! \mathcal{C}}$:
\begin{align}
\begin{split}
\label{set:Ce}
\mathcal{C}_{E} \triangleq \Bigg\{x \in \mathcal{C} ~\Bigg| \begin{array}{ll} 
\overbrace{h (\phi_{\rm b}(\tau,x))}^{\triangleq \bar{h}_{\tau}(x)} \geq 0, ~\forall \tau \in[0, T], \\
\underbrace{h_{\rm b}( \phi_{\rm b}(T,x))}_{\triangleq \bar{h}_{\rm b}(x)} \geq 0
\end{array}\Bigg\} .
\end{split}
\end{align}
That is, $\mathcal{C}_{E}$ is the set of initial states from where the system can use a $T$-length feasible controlled trajectory (that satisfies the input constraints and respects the system dynamics) to safely reach $\mathcal{C}_{\rm b}$.
We note that the input limits are satisfied since they are incorporated into the set ${\mathcal{C}_{E}}$ via ${\mathbf{k_{b}}}$. In~\eqref{set:Ce}, the first constraint implies that the flow under the backup controller satisfies the safety constraints, and the second constraint enforces that the backup set is reached in time ${T}$. To guarantee safety with respect to $\mathcal{C}_{E}$, we enforce the following constraint for all ${x \in \mathcal{C}_{E}}$:
\begin{align}
\small
\begin{split}
\label{eq:bcf_fin}
   \overbrace{ \dfrac{\partial h (\phi_{\rm b}(\tau,x))}{ \partial \phi_{\rm b}(\tau,x)}  \dfrac{\partial \phi_{\rm b}(\tau,x)}{\partial x}  f(x) }^{\triangleq L_f \bar{h}_{\tau}(x)} \!+\! \overbrace{ \dfrac{\partial h (\phi_{\rm b}(\tau,x))}{ \partial \phi_{\rm b}(\tau,x)}  \dfrac{\partial \phi_{\rm b}(\tau,x)}{\partial x}  g(x) }^{\triangleq L_g \bar{h}_{\tau}(x)} \!u \\
   \geq -\alpha (h (\phi_{\rm b}(\tau,x))), \quad \forall \tau \in[0, T], \\
\underbrace{ \dfrac{\partial h_{\rm b} (\phi_{\rm b}(T,x))}{ \partial \phi_{\rm b}(T,x)}  \dfrac{\partial \phi_{\rm b}(T,x)}{\partial x}  f(x) }_{\triangleq L_f \bar{h}_{\rm b}(x)} \!+\! \underbrace{ \dfrac{\partial h_{\rm b} (\phi_{\rm b}(T,x))}{\partial \phi_{\rm b}(T,x)}  \dfrac{\partial \phi_{\rm b}(T,x)}{\partial x}  g(x) }_{\triangleq L_g \bar{h}_{\rm b}(x)} \!u \\
\geq -\alpha_{\rm b} (h_{\rm b} (\phi_{\rm b}(T,x))) .
\end{split}
\end{align}
\begin{theorem}\cite{gurriet2020}
\label{theo:bcbf}
Any Lipschitz continuous controller that satisfies \eqref{eq:bcf_fin} keeps the closed loop system \eqref{eq:clsystem1} safe with respect to $\mathcal{C}_{E}$, which implies ${x(t) \!\in\! \mathcal{C}_{E} \!\subseteq\! \mathcal{C}, \forall t \!\geq\! 0}$ if ${x_0 \!\in\! \mathcal{C}_{E}}$.
\end{theorem}
Note that enforcing the first constraints in \eqref{eq:bcf_fin} is not computationally tractable, as it must hold for all ${\tau \in[0, T]}$. The constraint can be discretized to a finite collection of constraints and then can be directly used for controller synthesis via the following quadratic program (BCBF-QP):
\begin{align}
\begin{array}{lll}
{\mathbf{{k}^*}(x)= \ }
\displaystyle  \argmin_{{u}  \in U} \ \ {\|{u} -\mathbf{k_d}(x) \|^2}  \\ [3mm]
~~~~~~~~~\textrm{s.t.} \ \ \ L_{{f}} \bar{h}_{\tau_{i}}({x})\!+\!L_{{g}} \bar{h}_{\tau_{i}}({x}) {u}  \!\geq\!-\alpha(h_{\tau_{i}}({x} )) \\ [2mm]
~~~~~~~~~~~~~~~~L_{{f}} \bar{h}_{\rm b}({x})\!+\!L_{{g}} \bar{h}_{\rm b}({x}) {u}  \!\geq\!-\alpha_{\rm b}(h_{\rm b}({x})),
\end{array}
\end{align}
for all ${\tau_{i} = i T / N_{\tau}, ~ i = 0, 1, \ldots , N_{\tau}}$, where ${N_{\tau} \in \mathbb{N} }$ is the number of constraints, and ${\mathbf{k_d} (x): \mathbb{R}^n \to U}$ is a desired controller. In the upcoming experiments, the desired controller is given by the human operator's vehicle velocity commands. 

\subsection{Deep Neural Networks}
We consider an $N_L$-layer deep neural network (DNN), ${\Pi: \mathbb{R}^{n_{x}} \rightarrow \mathbb{R}^{n_{y}}}$, which is a piecewise linear function and maps an input feature vector ${\mathbf{\bar{x}} \triangleq (\mathbf{\bar{x}}_1, \hdots, \mathbf{\bar{x}}_{n_x}) \in \mathbb{R}^{n_x}}$ to an output vector ${\mathbf{\bar{y}} \triangleq (\mathbf{\bar{y}}_1, \hdots, \mathbf{\bar{y}}_{n_y}) \in \mathbb{R}^{n_y}}$, given by
\begin{align*}
\begin{split}
&\upsilon_{\jmath} =\pi_{\jmath}(\mathbf{\bar{W}}_{\jmath} \upsilon_{\jmath-1}+b_{\jmath}), \quad \jmath=1, \ldots, N_L, \\
&\Pi(\mathbf{\bar{x}}) =\mathbf{\bar{y}},
\end{split}
\end{align*}
owhere $\upsilon_{\jmath}$ is the output of the $\jmath$-th layer of the DNN, and ${\upsilon_{0} = \mathbf{\bar{x}}}$ is the input to the DNN and ${\upsilon_{N_L} = \mathbf{\bar{y}}}$ is the output of the DNN, respectively. Each hidden layer has ${n_{\jmath} \in \mathbb{N}}$ hidden neurons. ${\mathbf{\bar{W}}_{\jmath} \in \mathbb{R}^{n_{\jmath} \times n_{\jmath-1}}}$ and ${b_{\jmath} \in \mathbb{R}^{n_{\jmath}}}$ are weight matrices and bias vectors for the $\jmath^{{th}}$ layer, respectively. ${\pi_{\jmath} \triangleq \left[\varrho_{\jmath}, \cdots, \varrho_{\jmath}\right]}$ is the concentrated activation functions of the $\jmath^{{th}}$ layer wherein ${\varrho_{\jmath}: \mathbb{R} \rightarrow \mathbb{R}}$ is the activation function, such as sigmoid or ReLU:
\begin{align*}
\varrho_{\textrm{sigmoid}}(\mathbf{\bar{x}}_j) \triangleq \frac{1}{1+e^{-\mathbf{\bar{x}}_j}}, \ \ \varrho_{\textrm{ReLU}}(\mathbf{\bar{x}}_j) \triangleq \max(0, \mathbf{\bar{x}}_j). 
\end{align*}
\section{Framework}
\label{sec:theory}
For organizational purposes, the following sections that describe our framework use the following definitions
\begin{itemize}
    \item[--] $\K$ is a set of ${m_k \!\in\! \mathbb{N}}$ backup controllers such that controller ${\mathbf{k_{bi}} \!\in\! \K}$ renders a backup set ${\mathcal{C}_{{\rm b}i} \!\subseteq\! \mathcal{C}}$ forward invariant

    \item[--] ${\kappa \!:\! \mathbb{R}_{\geq 0} \to \mathbb{Z}}$ is a function that identifies the index of the chosen backup strategy as a function of time.
    
    \item[--] ${\gamma_{t, \kappa(t)} \!\in\! \mathbb{R}^k}$ is a vector of ${k \!\in\! \mathbb{N}}$ features--robot state, environment, driver input, and safety data--collected while the BCBF framework utilized controller ${\mathbf{k_{b \kappa(t)}}}$. 
    
    \item[--] We maintain an ${H\!+\!1}$-length data history ${\vec{\Gamma}(t, H, \kappa) \!=\! \left(\gamma_{t, \kappa(t)}, \gamma_{t-1, \kappa(t-1)}, \hdots, \gamma_{t-H, \kappa(t - H)}\right)}$. See that the chosen backup controller may change over the length of the history; it may also remain the same (for instance, $\kappa(t)$ may or may not equal $\kappa(t-1)$). Additional details on  $\vec{\Gamma}(t, H, \kappa)$ are found in the following discussion.
    \item[--] $R : \mathbb{R}^{k\times (H+1)} \to \mathbb{R}^{m_k}$ is a reward function that maps input history $\vec{\Gamma}(t, H, \kappa)$ to a list of $m_k$ rewards with values ranging from 0 to 1. 
\end{itemize}
We use a unicycle model to capture the tracked robot's motion, given by
\begin{equation}
\label{eq:uni_nom}
    \underbrace{
    \begin{bmatrix}
    \dot{x}_I \\
    \dot{y}_I \\
     \dot{\theta } 
     \end{bmatrix}}_{\dot{x}}
     = 
     \underbrace{\begin{bmatrix}
    0 \\
    0 \\
     0 
     \end{bmatrix}}_{{f}(x)} +
     \underbrace{
     \begin{bmatrix}
    \cos{\theta} & 0    \\
     \sin{\theta} & 0  \\
    0 & 1 
    \end{bmatrix}}_{{g}(x)}
    \underbrace{
    \begin{bmatrix}
        v \\ \omega
    \end{bmatrix}}_{u} ,
\end{equation}
where ${p \!=\! [ {x}_I ~ {y}_I ]^\top}$ is the vehicle's planar position w.r.t.~the inertial frame ${I}$, $\theta$ is vehicle's yaw angle, ${-v_{\rm max} \!\leq\! v \!\leq \!v_{\rm max}}$ is its linear velocity and ${-\omega_{\rm max} \!\leq\! \omega \leq \!\omega_{\rm max}}$ is the angular velocity. While we describe our intention estimator in the context of the unicycle model, we believe that our system can generalize to more complex dynamical models. This is the subject of future experimentation as detailed in Section~\ref{sec:future}.

\begin{figure}[t]
    \centering
    \includegraphics[width=8.6cm]{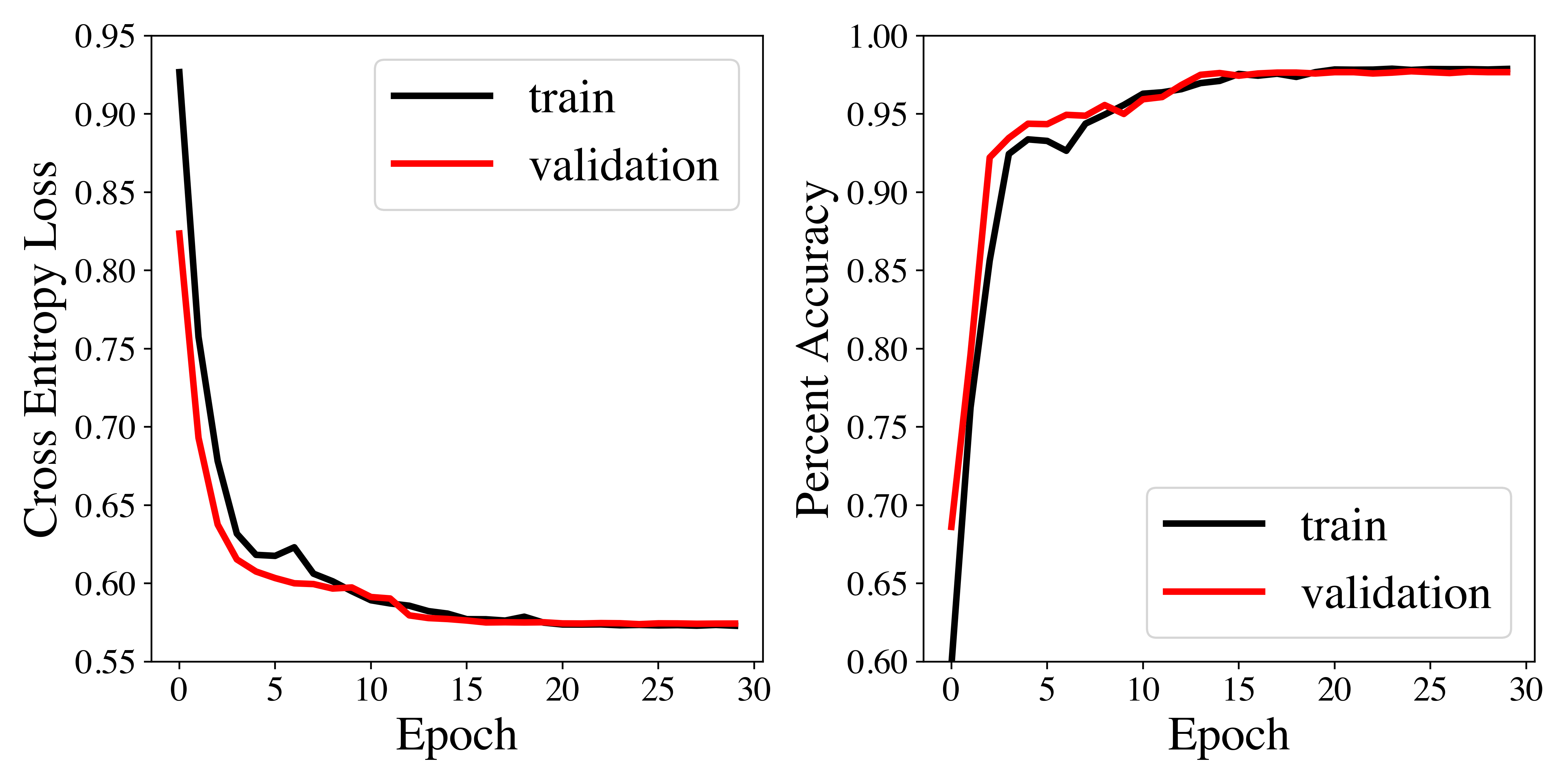}
    \vskip -0.05 true in
    \caption{Cross entropy and training loss during a training episode. Accuracy is calculated by comparing the maximum reward ouput from the LSTM-DNN architecture and determining if it corresponds to the correct choice of backup controller for the respective point in the training set. We also use a validation (test) dataset to observe out-of-sample performance of the network during training. The network achieves 97\% accuracy by the 30th epoch. }
    \label{fig:training}
    \vskip -0.2 true in
\end{figure}

\subsection{Intention Estimation}

We employ an intention estimation framework to learn the reward function, $R$. Learning is useful in this context since constructing a reward function based on multi-modal input in high dimensions may be challenging. The feature vector at a single timestep, ${\gamma_{t, \kappa(t)}}$, consists of the robot position ${[x_I, y_I, z_I, \theta] \in \mathbb{R}^4}$, robot velocities ${[\dot{x}_I, \dot{y}_I, \dot{z}_I, \dot{\theta}] \in \mathbb{R}^4}$, driver velocity commands ${u \in U}$, and safety information from evaluating ${h: \mathbb{R}^3 \to \mathbb{R}}$ at $x$ and at ${x_g \in \mathbb{R}^3}$, which is an intermediate goal  determined by forward integrating the driver's desired $u$ over a short horizon. These quantities are measured and calculated while the BCBF safety filter wtih backup controller $\mathbf{k_{b \kappa(t)}}$ is influencing the robot trajectory. Hence, $\kappa(t)$ is designated to the feature vector $\gamma_{t,\kappa(t)}$ to indicate this dependence. Note that $\gamma_{t,\kappa(t)}$ can contain many more features than detailed here in other implementations, especially if they relate to the safety of the robot. However, the aforementioned features generalize to any robot using our implementation. We wish to learn $R$ by mapping history  $\vec{\Gamma}(t,H, \kappa)$ to rewards corresponding to each potential backup controller $\mathbf{k_{b\kappa(t+1)}}$ at next timestep $t+1$. This process allows us to derive switching laws for the \textit{next}  backup controller as a function of time. For instance, we may choose the backup controller with the largest reward at time $t+1$ where $\kappa(t+1) = \argmax(R(\vec{\Gamma}(t,H, \kappa))$. This recursive formulation necessitates that we initialize $\kappa(t_0)$ such that the robot begins in the safe control invariant set $\mathcal{C_E}$ corresponding to backup controller $\mathbf{k_{b\kappa(t_0)}}$

We use a deep LSTM network with a DNN decoder to construct $R$ from history $\vec{\gamma}(t, H, \kappa)$. Deep LSTMs can learn complex temporal relationships by using multiple layers and cyclic connections to understand sequential data, and the DNN decoder maps the output of the LSTM to rewards for each backup controller through several hidden layers. Both the LSTM and DNN utilize dropout for regularization, and the DNN uses ReLU activation functions in the hidden layers. Furthermore, we use the sigmoidal activation function in the DNN final layer. Thus, the outputted rewards for each backup controller, $\mathbf{k_{bi}}$, at time $t$ (where $i$ is the index of the output of $R$) can be interpreted very loosely as a likelihood that $\mathbf{k_{bi}}$ is the desired backup controller choice at time $t$. For example, if our framework is highly certain that the human operator 
believes $\mathbf{k_{b0}}$ is the correct choice of backup controller at time $t$, then the first component of the output of ${R(\vec{\gamma}(t, H, \kappa))}$ is expected to be close to 1.

Our strategy and architecture enables easy reward engineering for the training dataset. We construct a multi-class dataset by gathering data of the robot driving in suitable environments under the correct backup controllers. During data collection the robot should be operating with a BCBF safety filter as we plan to evaluate the network in safety-critical contexts. During training, the switches between backup controllers are labeled in the dataset by the driver using configurable buttons on the robot ground station. Labeling is completed by assigning $1$ to the driver's choice of backup controller and $0$ to all others for that instance. Since we utilize the sigmoidal activation function in this multi-class classification learning task, we use the softmax loss function to train the network. Indeed, this training procedure makes an implicit assumption that there exists a single, correct choice of backup controller at any given time. While this may not be true in certain situations, we guarantee safe switching as discussed in Section \ref{subsec:switching}. Thus, any safe switch between backup controllers, does not void safety guarantees. 

To improve training, we utilize the Adam optimizer \cite{kingma2014adam} with a stepped learning rate scheduler. In order to compensate for the time it takes to solve the BCBF-QP, we shift the labels for the desired backup controller backwards in time. This allows the predicted backup strategy to be preemptively passed into the BCBF-QP, allowing time for the BCBF-QP to have computed safe control outputs for the new controller when they are expected. 

We present the model parameters used in our final implementation and training iteration shown in Fig. \ref{fig:training}. We used a 2 layer LSTM with 100 hidden units with the 12 total input features detailed earlier. Our DNN is composed of two hidden layers of sizes 50 and 25 neurons, respectively. We use a dropout value of 0.2 for the hidden layers of the DNN and a dropout of 0.1 for the LSTM. Accuracy is calculated by comparing the maximum reward output from the LSTM-DNN architecture and determining if it corresponds to the correct choice of backup controller for the respective point in the training set. We also use a validation (test) dataset to observe out-of-sample performance of the network during training. The network achieves 97\% accuracy by the 30th epoch. 
We achieved this accuracy on a dataset of 19000 datapoints collected on hardware. The sequence length for the training samples of the model was chosen to be 15 timesteps, which corresponds to roughly 0.75 seconds of data. We found that this time-range produced the most accurate results. 

\subsection{Backup Control Barrier Function Modification} \label{subsec:switching}
Safety must be preserved when switching between backup controllers. Consider a switch from backup controller $\mathbf{k_{bi}}$ to $\mathbf{k_{bj}}$. One way to ensure safety during this crossover is to evaluate constraints  \eqref{eq:bcf_fin} for the new controller $\mathbf{k_{bj}}$. Since we were previously using backup strategy $\mathbf{k_{bi}}$, we are guaranteed to be in the $T$-time reachable set of the first controller. Thus, validating that the BCBF-QP constraints hold for $\mathbf{k_{bj}}$, practically requires that both backup controllers' $T$-time reachable sets intersect, and allows the BCBF-QP to remain solvable before and after the transition. As observed in our implementation ( \ref{sec:implementation}), this resulted in relatively smooth switching between backup strategies. 

\subsection{Human Interface}

Since we extract an intention estimation signal from the human driver, providing feedback is essential for improving human-robot collaboration. Past work has suggested that systems which allow robots to estimate human intention, while their collaborating humans estimate robot intention, can improve overall performance \cite{DraganIntention}. Thus, we develop an interface based on the ROS \textit{rviz} visualization tool as well as an Xbox controller featuring haptic motors. The \textit{rviz} system displays a visualization of the robot pose, along with dials that indicate estimated system safety, as quantified by the magnitude of the saturated safety function $\tanh(h(x))$. Furthermore, vibratory haptic feedback is provided to the user when the robot nears the boundary of the safe set (indicated with $h$). We tune the the thresholds and strength of the haptic feedback as preferred by the driver. 

\begin{figure}[]
    \centering
    \includegraphics[width=6.0cm]{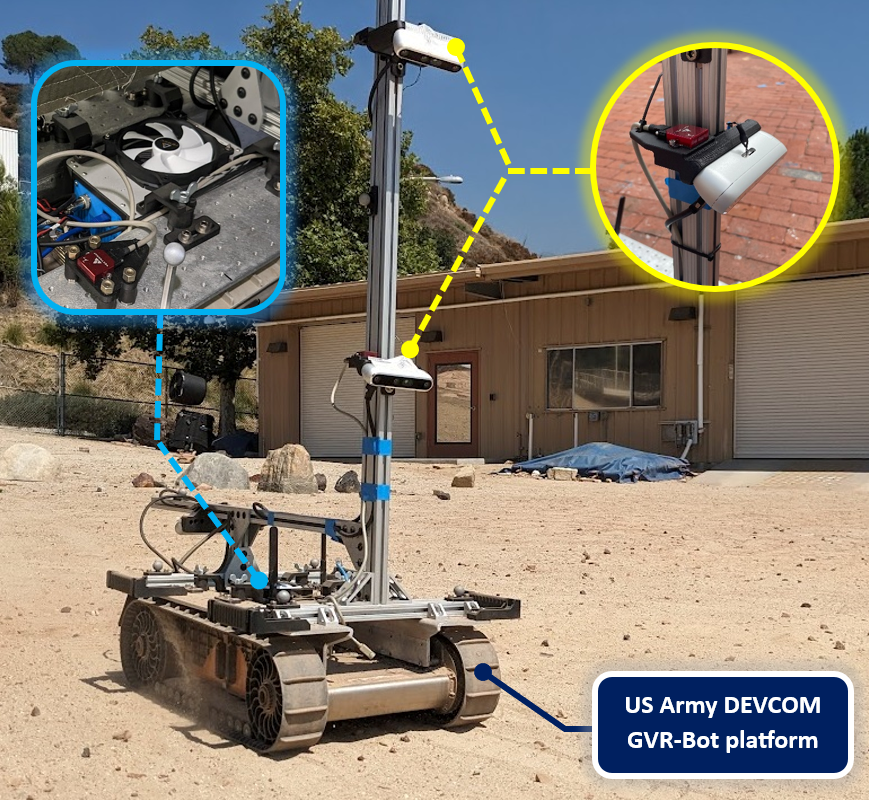}
    \caption{Our test-vehicle (US Army DEVCOM GVR-Bot robot) in the NASA Jet Propulsion Laboratory Mars Yard. $(yellow\ inset)$ Intel Realsense D457 Depth Cameras are coupled with a VN-100 IMU, $(light-blue\ inset)$ custom compute payload.}
    \label{fig:mbcbf_hw_final_1}
    \vskip -0.4 true in 
\end{figure}
\section{Implementation} \label{sec:implementation}
The aforementioned framework was deployed in a simple obstacle avoidance task in a 2D environment. The robot position is measured from its center, so a buffer radius is given to the circular obstacle (a traffic cone) to account for half the length of the robot. The robot is driven to various locations around the environment to assess the accuracy of our framework to predict the desired backup strategy.

\begin{figure*}[!th]
\label{fig:result_icra24}
    \centering
    \includegraphics[width=18cm]{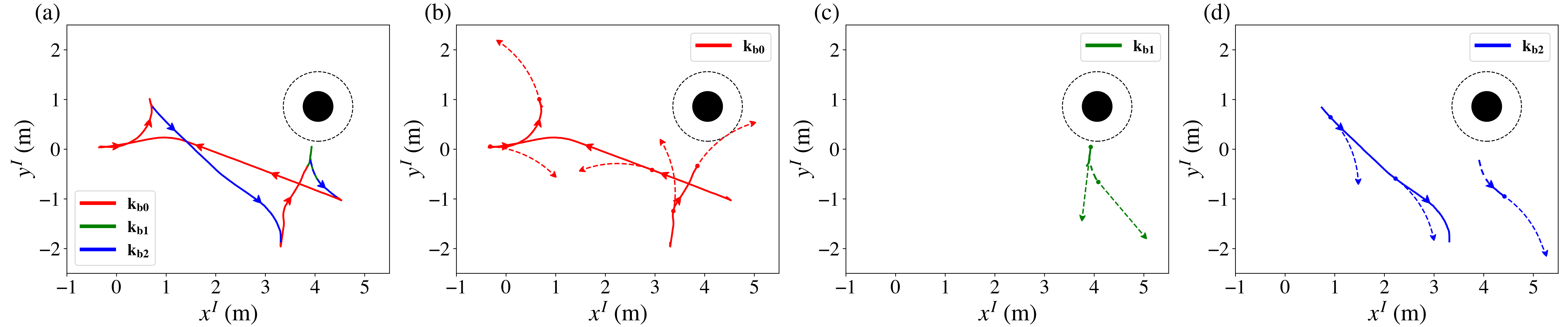}
    
    \caption{Robot trajectory which used all three backup controllers under our learned switching law. Subfigure (a) shows the robot trajectory, where direction is indicated by the arrows and the choice of backup controller is indicated by color. Subfigures (b), (c), and (d) show the segments of the robot trajectory that used $\mathbf{k_{b0}}, \mathbf{k_{b1}}, \mathbf{k_{b2}}$ respectively. In (b), (c), and (d) we also show the flows of the respective backup controllers at several robot positions. Notice that in each of (b) (c) (d), the backup controller flow always escapes the obstacle. However, notice that $\mathbf{k_{b0}}$ may not provide safe evasion from the obstacle in the locations in subfigure (c) as it does for the locations in (b). This implies that our system chose the correct backup controller depending on several factors, like driver intent and robot position.}
    \label{fig:result}
    \vskip -0.1 true in
\end{figure*} 
\subsection{Hardware}
Our algorithms are deployed on a tracked GVR-Bot from US Army DEVCOM Ground Vehicle Systems Center. The GVR-Bot is a modified iRobot PackBot 510 and its rugged design and quick actuation makes it ideal for research of safety in the presence of unknown human driver intentions. Our Python and C++ algorithms run on a custom compute payload that is based on an NVIDIA Jetson AGX Orin (2048-core NVIDIA Ampere architecture, 64 Tensor Cores, 275 TOPS, 12 Core Arm Cortex A78AE @ 2.2GHz, 32GB Ram). Vision is provided by three synchronized Intel Realsense D457 Depth Cameras which are strategically positioned to provide a wide field of view for the control system and operator. They operate at a 30 Hz frame rate. A Vectornav VN-100 provides inertial measurements. Communication between our algorithm, the control computer, and the internal GVR-Bot drive computer (Intel Atom) is facilitated via ROS1.  Estimation of the vehicle state is provided by an OpenVINS visual-inertial estimator \cite{genevaOpenVINS}. Drive commands (linear and angular body velocities) are communicated from the AGX Orin to the Intel Atom where they are converted into individual track speeds, which are regulated via high-rate controllers on the GVR-Bot, see Fig.~\ref{fig:master} and Fig.~\ref{fig:mbcbf_hw_final_1}.

\subsection{Backup Controllers}
\label{subsec:MBCBF}
For safety, the tracked robot must avoid moving obstacles, considered as cylinders with radius ${R_o \in \mathbb{R}_{> 0}}$, position
${p_o \in \R^2}$
and velocity ${v_o \in \R^2}$.
This leads to the safety constraint ${h(x) \geq 0}$, with the CBF candidate given by
\begin{equation*}
    h(x) = \|p - p_o\|-R_o.
\end{equation*}
We enforce safety in the presence of input bounds by implementing multiple BCBFs. We leverage three backup controllers that yield qualitatively different behavior.
Controller $\mathbf{k_{b0}}$ turns the robot away from the obstacle and drives forward.
Controller $\mathbf{k_{b1}}$ drives straight away as the obstacle is approached, without turning.
Controller $\mathbf{k_{b2}}$ turns towards the obstacle and drives in reverse. It behaves similarly to $\mathbf{k_{b0}}$, however the robot has turned around.
These are expressed by:
\begin{align*}
    & \mathbf{k_{b}}(x) = \begin{bmatrix}
        v_{\rm max} \\
        \omega_{\rm max} \tanh(n^Tr/\varepsilon)
    \end{bmatrix}, \;\;
    h_0(x) = n^T(q v_{\rm max} - v_o), \\
    & \mathbf{k_{b1}}(x) = \begin{bmatrix}
        v_{\rm max}\tanh(n^Tq/\varepsilon) \\
        0
    \end{bmatrix}, \;\;
    h_1(x) = \|p - p_o\|-R_o, \\
    & \mathbf{k_{b2}}(x) = - \mathbf{k_{b0}}(x), \hspace{20mm}
    h_2(x) = -h_0(x),
\end{align*}
where $\tanh$ is used to obtain smooth policies with smoothing parameter ${\varepsilon \in \R_{>0}}$, and the vectors $n$, $q$, $r$ are given by:
\begin{equation}
    n = \frac{p - p_o}{\| p - p_o \|}, \quad
    q = \begin{bmatrix}
    \cos \theta \\ \sin \theta
    \end{bmatrix}, \quad
    r = \begin{bmatrix}
    -\sin \theta \\ \cos \theta
    \end{bmatrix}.
\end{equation}
Each policy maintains a forward invariant backup set, i.e., the 0-superlevel set of the BCBF $h_0$, $h_1$ and $h_2$. These sets represent the states where the robot faces away from the obstacle (for $h_0$), the robot has positive distance from the obstacle ($h_1$), and it faces towards the obstacle ($h_2$).

Ultimately, the BCBFs enable the robot to maintain safe behavior.
At the same time, the performance can be improved by switching between the backup policies based on learning.

\section{Results}

\label{sec:results} 
\begin{figure}
    \centering
    \includegraphics[width=8.6cm]{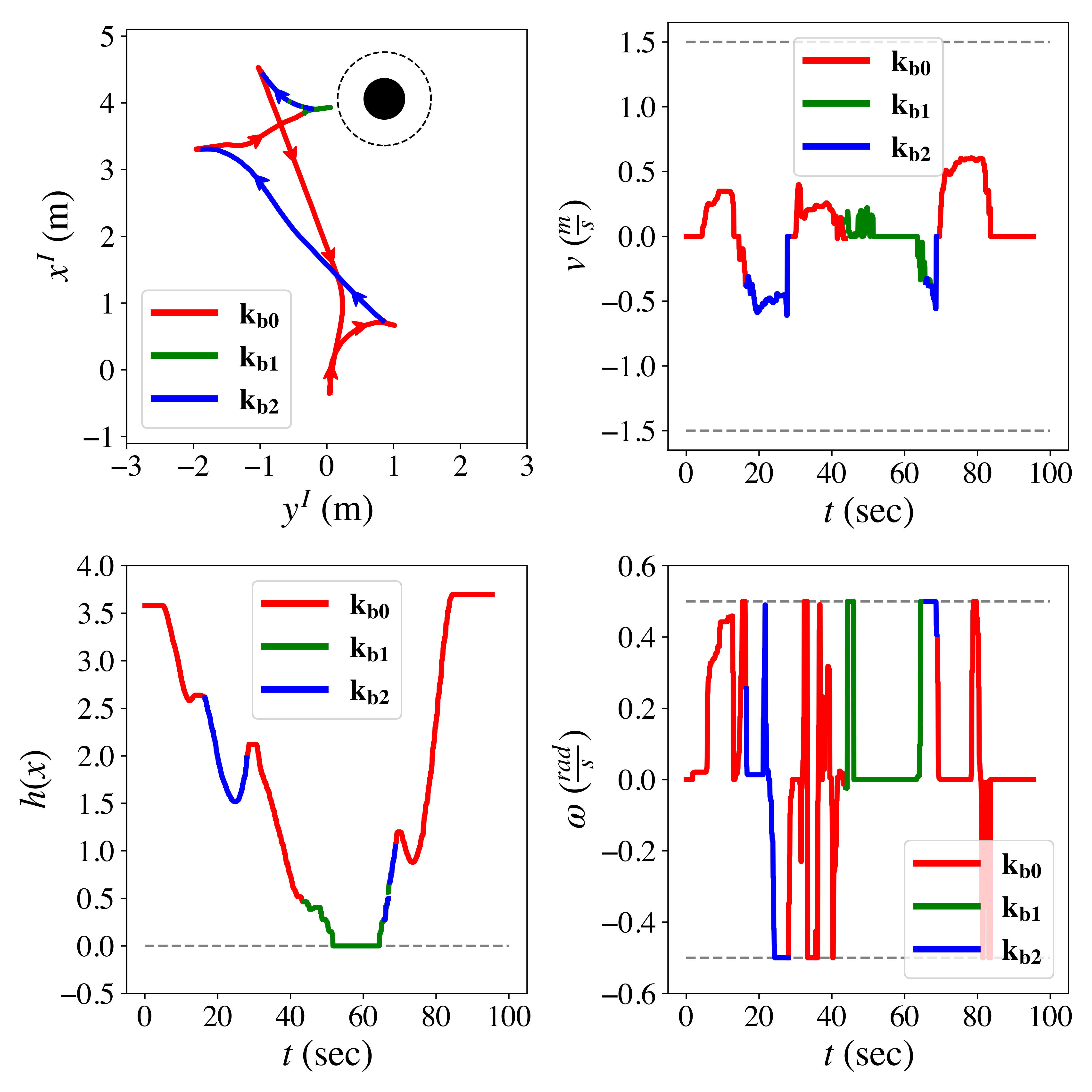}
     \vskip -0.05 true in
    \caption{Demonstration of system safety during switches between backup controllers. See that $h(x) \geq 0$ and the input constraints are satisfied (denoted by the gray dashed lines); therefore, our system maintains safety while better aligning with driver intention.}
    \label{fig:safety}
    \vskip -0.2 true in
\end{figure}

By referring to Fig.~\ref{fig:result}, it is evident from the alternating colors that our learned reward function selected multiple backup controllers over a single robot trajectory.  Our framework also selected appropriate backup controllers such that robot reachability near the obstacle was maximized.  Furthermore, our system maintains the formal safety guarantees from the BCBF method as demonstrated in Fig.~\ref{fig:safety}, where $h$ was observed to be positive under input limits. To demonstrate the reproducibility of these results, further tests were conducted and documented in the accompanying video, shared in the caption of Fig. \ref{fig:master}. Here, we tested our system in many scenarios and compared our system's produced trajectories to trajectories corresponding to a single backup controller. The use of our switching framework outperforms the use of a single backup controller in certain scenarios, and our system never inhibits the driver from achieving their goal, while some backup controllers do. 

\section{Conclusions and Future Work} \label{sec:future}
In this work, we implemented backup control barrier functions to ensure safety on a tracked robot platform, and used driver intention estimation to optimally choose between multiple backup policies.
From our results, we conclude that this system improves safety and performance by improving the interaction between the robot and its driver by expanding the multiple BCBF framework with a learning method. 

Several next steps exist for this preliminary framework, like developing an algorithm to formally compute safe reachability under multiple backup controllers. Furthermore, while our framework was demonstrated on a teleoperated ground vehicle with stationary obstacles, we plan to deploy our framework on other robots, such as quadrupeds or drones, with moving obstacles.
Finally, various techniques can be employed to enhance the resilience of standard CBF constraint \eqref{eq:cbf_def} against disturbance or model uncertainty, such as GP-based uncertainty in the CBF constraint \cite{castaneda2021}, \cite{castaneda2021pointwise}. 

\vskip 0.07 true in
\textbf{Acknowledgment.} The authors would like to thank the SFP program at Caltech, sponsors Kiyo and Eiko Tomiyasu, and DARPA 
for funding this project. The authors would also like to thank Ryan Cosner and Maegan Tucker for discussions about BCBFs and learning, and Matthew Anderson for his insights and help in experimental setup and testing.

\clearpage
\bibliographystyle{IEEEtran}
\bibliography{References.bib}

\end{document}